\documentclass[]{spie}  

\usepackage{subcaption}
\usepackage{setspace}

\usepackage{amsmath,amsfonts,amssymb}
\usepackage{graphicx}
\usepackage[colorlinks=true, allcolors=blue]{hyperref}
\usepackage[leftcaption]{sidecap}

\title{Large-Scale Extended Granger Causality for Classification of Marijuana Users From Functional MRI}

\author[a]{M. Ali Vosoughi}
\author[a,b,c,d]{Axel Wismüller}

\affil[a]{Department of Electrical and Computer Engineering, University of Rochester, NY, USA}
\affil[b]{Department of Imaging Sciences, University of Rochester, NY, USA}
\affil[c]{Department of Biomedical Engineering, University of Rochester, NY, USA}
\affil[d]{Faculty of Medicine and Institute of Clinical Radiology, Ludwig Maximilian University,
Munich, Germany}

\authorinfo{Further author information: (Send correspondence to Ali Vosoughi)\\Ali Vosoughi: E-mail: mvosough@ur.rochester.edu}

\begin{document} 
\maketitle

\begin{abstract}
It has been shown in the literature that marijuana use is associated with changes in brain network connectivity. We propose large-scale Extended Granger Causality (lsXGC) and investigate whether it can capture such changes using resting-state fMRI. This method combines dimension reduction with source time-series augmentation and uses predictive time-series modeling for estimating directed causal relationships among fMRI time-series. It is a multivariate approach, since it is capable of identifying the interdependence of time-series in the presence of all other time-series of the underlying dynamic system. Here, we investigate whether this model can serve as a biomarker for classifying marijuana users from typical controls using 126 adult subjects with a childhood diagnosis of ADHD form the Addiction Connectome Preprocessed Initiative (ACPI) database. We use brain connections estimated by lsXGC as features for classification. After feature extraction, we perform feature selection by Kendall's-tau rank correlation coefficient followed by classification using a support vector machine. As a reference method, we compare our results with cross-correlation, which is typically used in the literature as a standard measure of functional connectivity. Within a cross-validation scheme of 100 different training/test (90\%/10\%) data splits, we obtain a mean accuracy range of [0.714, 0.985] and a mean Area Under the receiver operating characteristic Curve (AUC) range of [0.779, 0.999] across all tested numbers of features for lsXGC, which is significantly better than results obtained with cross-correlation, namely mean accuracy of [0.728, 0.912] and mean AUC of [0.825, 0.969]. Our results suggest the applicability of lsXGC as a potential biomarker for marijuana use.

\end{abstract}


\section{INTRODUCTION} \label{sec:intro}  
Marijuana has been used as an illegal drug for decades. Despite many concerns about the adverse effects of repeated marijuana abuse on brain functionalities and cognition, considerably few studies have been conducted to investigate the effects of repeated marijuana use on the brain [\citeonline{solowij2006cannabis}]. Currently, the diagnosis of cannabis use disorder is based on the clinical evaluations of the symptoms and behaviors.  However, more objective biomarkers are of interest. To this end, more recently, studies have investigated, if information can be extracted non-invasively from brain activity data.  Although these studies have shown promising results, there is still scope for improvement, especially with regards to using more meaningful connectivity analysis approaches [\citeonline{dadi2019benchmarking}].

Some evidence has demonstrated that frequent marijuana use affects the connectivity of the brain [\citeonline{dadi2019benchmarking, solowij2006cannabis}].  Biomarkers from resting-state functional MRI (rs-fMRI) for frequent marijuana use can be derived using Multi-Voxel Pattern Analysis (MVPA) techniques [\citeonline{4_norman2006beyond,dadi2019benchmarking}]. MVPA is a machine-learning framework that extracts differences in patterns of brain connectivity to discriminate between connectivity profiles of individuals with neurological diseases/disorders and healthy individuals. Most MVPA studies commonly use cross-correlation to construct a functional connectivity profile. For example, one such study has obtained an accuracy of 0.69 on the slow frequency bands (0.01-0.1 Hz) [\citeonline{dadi2019benchmarking}]. These results demonstrate that meaningful information can be learned from fMRI data. However, correlation does not provide a measure of directed connectivity. Hence, there may be more relevant information in the fMRI data that is not being captured by cross-correlation.

Various methods have been proposed to obtain directional relationships in multivariate time-series data, \textit{e.g.}, transfer entropy [\citeonline{schreiber2000measuring}] and mutual information [\citeonline{kraskov2004estimating}]. However, as the multivariate problem's dimensions increase, computation of the density function becomes computationally expensive [\citeonline{mozaffari2019online,mozaffari2019online_ieee}]. Under the Gaussian assumption, transfer entropy is equivalent to Granger causality [\citeonline{barnett2009granger}]. However, the computation of multivariate Granger causality for short time series in large-scale problems is challenging [\citeonline{vosoughi2020large,dsouza2020large}]. 
We introduce large-scale Extended Granger Causality (lsXGC) as a novel method that combines dimension reduction with source time-series augmentation and uses predictive time-series modeling for estimating directed causal relationships among fMRI time-series [\citeonline{wismuller2020large}]. 
In this work, we investigate, if changes in directed connectivity manifest the frequent marijuana use among adults with a childhood diagnosis of ADHD and if such directed measures are better able to discriminate between frequent marijuana users and typical controls. To this end, we apply lsXGC in the MVPA framework for estimating a measure of directed causal interdependence between fMRI time-series.

This work is embedded in our group’s endeavor to expedite artificial intelligence in biomedical imaging by means of advanced pattern recognition and machine learning methods for computational radiology and radiomics, \textit{e.g.} [ \citeonline{nattkemper2005tumor,bunte2010adaptive,8_wismueller2000segmentation,9_leinsinger2006cluster,10_wismuller2004fully,11_hoole2000analysis,12_wismuller2006exploratory,13_wismuller1998neural,14_wismuller2002deformable,15_behrends2003segmentation,16_wismuller1997neural,17_bunte2010exploratory,18_wismuller1998deformable,19_wismuller2009exploration,20_wismuller2009method,22_huber2010classification,23_wismuller2009exploration,24_bunte2011neighbor,25_meyer2004model,26_wismuller2009computational,27_meyer2003topographic,28_meyer2009small,29_wismueller2010model,meyer2007unsupervised,30_huber2011performance,31_wismuller2010recent,meyer2007analysis,32_wismueller2008human,wismuller2015method,33_huber2012texture,34_wismuller2005cluster,35_twellmann2004detection,37_otto2003model,38_varini2004breast,39_huber2011prediction,40_meyer2004stability,41_meyer2008computer,42_wismuller2006segmentation,45_bhole20143d,46_nagarajan2013computer,47_wismuller2004model,48_meyer2004computer,49_nagarajan2014computer,50_nagarajan2014classification,yang2014improving,wismuller2014pair,51_wismuller2014framework,schmidt2014impact,wismuller2015nonlinear,wismuller2016mutual,52_schmidt2016multivariate,abidin2017using,61_dsouza2017exploring,53_chen2018mri,54_abidin2018alteration,55_abidin2018deep,dsouza2018mutual,chockanathan2019automated} ].
\section{DATA} \label{sec:data}
This data respiratory of the Multimodal Treatment Study of ADHD (MTA) contains a longitudinal study of participants for fourteen years pursued at multiple sites in North America. The study is one of the largest conducted studies for ADHD diagnosis and treatment, with primary results being published in 1999. The Addiction Connectome Preprocessed Initiative (ACPI) data respiratory contains datasets from a subset of MTA participants at six sites with and without childhood ADHD, who were studied as part of a follow-up multimodal brain imaging examination. The principal aim of the MTA study was to investigate the effect of cannabis use among adults with a childhood diagnosis of ADHD. The study was a 2 x 2 design of those with and without childhood ADHD and those who did or did not regularly use cannabis [\citeonline{MTA2015.}].

The scan parameters of resting-state fMRI the ACPI dataset can be found on the website of the project [\citeonline{MTA2015.}]. Preprocessed data from the MTA 1 dataset [\citeonline{MTA2015.}] of the ACPI database was used. The dataset included 126 subjects, 101 males and 25 females, with ages between 21-27 years, 86 diagnosed with ADHD (68\%), and 62 of them regularly used marijuana (49\%). Preprocessing of the raw 4D rs-fMRI data had been made using a Configurable Pipeline for the Analysis of Connectomes (C-PAC) [\citeonline{C-PAC}] and the Advanced Normalization Tools (ANTs) pipeline and consisted in the removal of the first five fMRI volumes, anatomical registration, tissue segmentation, functional registration in the Montreal Neurological Institute (MNI) space, functional masking, temporal bandpass filtering (0.01 - 0.1 Hz), motion correction, spatial smoothing, and various nuisance corrections [\citeonline{indi-team}]. The MODL parcellation has been used as a pre-computed atlas for regions definition and built using a form of online dictionary learning [\citeonline{dadi2019benchmarking}]. 
\section{METHODS}\label{sec:methods}
\subsection{Large-scale Extended Granger Causality (lsXGC)}

Large-scale Extended Granger Causality (lsXGC) has been developed based on 1) the principle of original Granger
causality, which quantifies the causal influence of time-series $\bold{x_s}$ on time-series $\bold{x_t}$ by quantifying the amount of improvement in the prediction of $\bold{x_t}$ in presence of $\bold{x_s}$. 2) the idea of dimension reduction, which resolves the problem of the tackling a under-determined system, which is frequently faced in fMRI analysis, since the number of acquired temporal samples usually is not sufficient for estimating the model parameters [\citeonline{vosoughi2020large,dsouza2020large}].

Consider the ensemble of time-series $\mathcal{X}\in \mathbb{R}^{N\times T}$, where $N$ is the number of time-series (Regions Of Interest – ROIs) and $T$ the number of temporal samples. Let $\mathcal{X} = (\bold{x_1}, \bold{x_2}, \dots, \bold{x_N})^{\mathsf{T}}$ be the whole multidimensional system and $x_i \in \mathbb{R}^{1\times T}$ a single time-series with $i = 1, 2, \dots,N$, where $\bold{x_i} = (x_i(1), x_i(2), \dots, x_i(T))$. In order to overcome the under-determined problem, first $\mathcal{X}$ will be decomposed into its first $p$ high-variance principal components
$\mathcal{Z} \in \mathbb{R}^{p\times T}$ using Principal Component Analysis (PCA), \textit{i.e.},

\begin{equation}
\mathcal{Z}=W\mathcal{X},    
\end{equation}

where $W\in \mathbb{R}^{p\times N}$ represents the PCA coefficient matrix. Subsequently, the dimension-reduced time-series ensemble $\mathcal{Z}$ is augmented by one original time-series $\bold{x_s}$ yielding a dimension-reduced augmented time-series ensemble $\mathcal{Y}\in \mathbb{R}^{(p+1)\times T}$ for estimating the influence of $\bold{x_s}$ on all other time-series.

Following this, we locally predict $\mathcal{X}$ at each time sample $t$, \textit{i.e.} $\mathcal{X}(t)\in \mathbb{R}^{N\times 1}$ by calculating an estimate $\hat{\mathcal{X}}_{\bold{x_s}}(t)$. To this end, we fit an affine model based on a vector of $m$ vector of m time samples of $\mathcal{Y}(\tau)\in \mathbb{R}^{(p+1)\times 1}$($\tau=t-1, t-2, \dots, t-m$), which is $\bold{y}(t)\in \mathbb{R}^{m.(p+1)\times 1}$, and a parameter matrix $\mathcal{A}\in \mathbb{R}^{N\times m.(p+1)}$ and a constant bias vector $\bold{b}\in \mathbb{R}^{N\times 1}$, 
\begin{equation}
    \hat{\mathcal{X}}_{\bold{x_s}}(t)=\mathcal{A}\bold{y}(t)+\bold{b},~~ t=m+1, m+2, \dots, T.
\end{equation}

Now $\hat{\mathcal{X}}_{\setminus {\bold{x_s}}}(t)$, which is the prediction of $\mathcal{X}(t)$ without the information of $\bold{x_s}$, will be estimated. The estimation processes is identical to the previous one, with the only difference being that we have to remove the augmented time-series $\bold{x_s}$ and its corresponding column in the PCA coefficient matrix $W$.

The computation of a lsXGC index is based on comparing the variance of the prediction errors obtained with
and without consideration of $\bold{x_s}$. The lsXGC index $f_{\bold{x_s}\xrightarrow{}\bold{x_t}}$ , which indicates the influence of $\bold{x_s}$ on $\bold{x_t}$, can be calculated by the following equation:
\begin{equation}
    f_{\bold{x_s}\xrightarrow{}\bold{x_t}}=\log {\frac{\mathrm{var}(e_s)}{\mathrm{var}(e_{\setminus s})}},
\end{equation}
where $e_{\setminus s}$ is the error in predicting $\bold{x_t}$ when $\bold{x_s}$ was not considered, and $e_s$ is the error, when $\bold{x_s}$ was used. In this study, we set $p = 8$ and $m = 4$.

\subsection{Multi-voxel pattern analysis}
In this study, brain connections served as features for classification and were estimated by two methods, namely lsXGC and cross-correlation. Before using high-dimensional connectivity feature vectors as input to a classifier, feature selection was carried out to reduce the dimension of input features.

\subsubsection{Feature selection}
To reduce the number of features, feature selection was performed on each training data set with k-fold cross-validation using \textit{Kendall’s Tau} rank correlation coefficient [\citeonline{62_kendall1945treatment}] and $10\%-90\%$ of test-to-train split ratio. This approach quantifies the relevance of each feature to the task of classification and assigns ranks by testing for independence between different classes for each feature [\citeonline{62_kendall1945treatment}]. 

\subsubsection{Classification}
For the 100 iteration cross-validation scheme for classification, the data set was divided into two groups: a training data set ($90\%$) and a test data set ($10\%$) in a way that the percentage of samples for each class was preserved.
Also, this was repeated with different numbers of features ranging from 5 to 175. A Support Vector Machine (SVM) [\citeonline{63_suykens1999least}] was used for classification between control subjects and marijuana users. All procedures were implemented using MATLAB 9.8 (MathWorks Inc., Natick, MA, 2020a) and Python 3.8.

\section{RESULTS}\label{sec:results}
\begin{SCfigure}[][!h]
    \centering
    \includegraphics[scale=0.39]{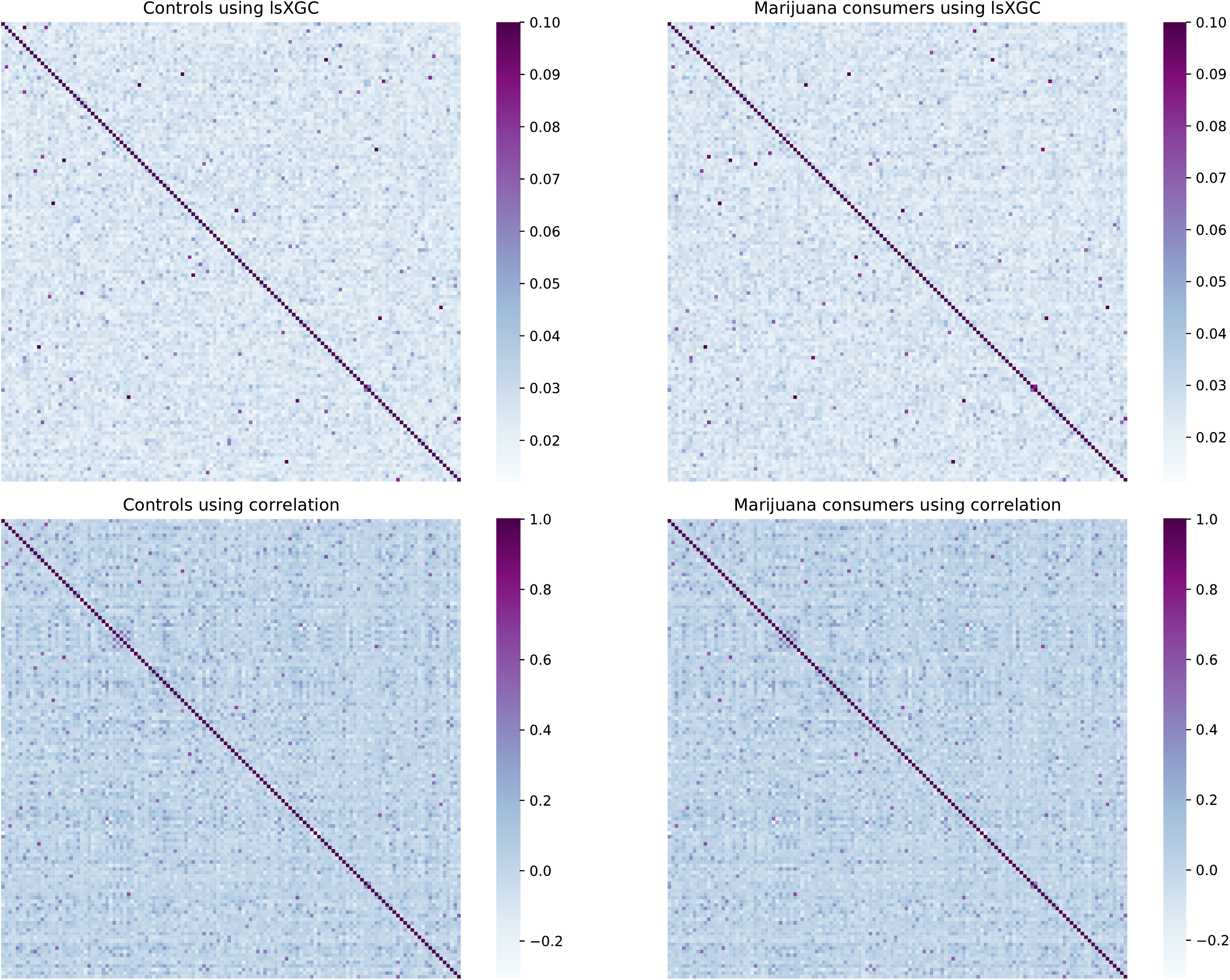}
    \caption{Mean connectivity matrices: top left: mean connectivity matrix of typical control subjects using lsXGC, top right: mean connectivity matrix of frequent marijuana users using lsXGC, bottom left: mean connectivity matrix of typical controls using cross-correlation, bottom right: mean connectivity matrix of frequent marijuana users using cross-correlation. Note that the different methods appear to extract different connectivity features and that they appear to be slight differences in connectivity patterns between the typical controls and the frequent marijuana users. }
    \label{fig:matrix_plot}
\end{SCfigure}

Mean connectivity matrices that were extracted using lsXGC and cross-correlation, are shown in Fig. \ref{fig:matrix_plot} for both typical control and frequent marijuana user cohorts. Different patterns are visible to the naked eye for both methods. In the following, we quantitatively investigate the difference between connectivity patterns of the two subject cohorts using an MVPA approach.

\vspace{-.1in}

Classification results were evaluated using the Area Under the Receiver Operator Characteristic Curve (AUC) and accuracy. An AUC = 1 indicates a perfect classification, AUC = 0.5 indicates random class assignment.
\begin{figure}
     
     \centering
     \begin{subfigure}[b]{0.4\textwidth}
         \centering
         \includegraphics[width=\textwidth]{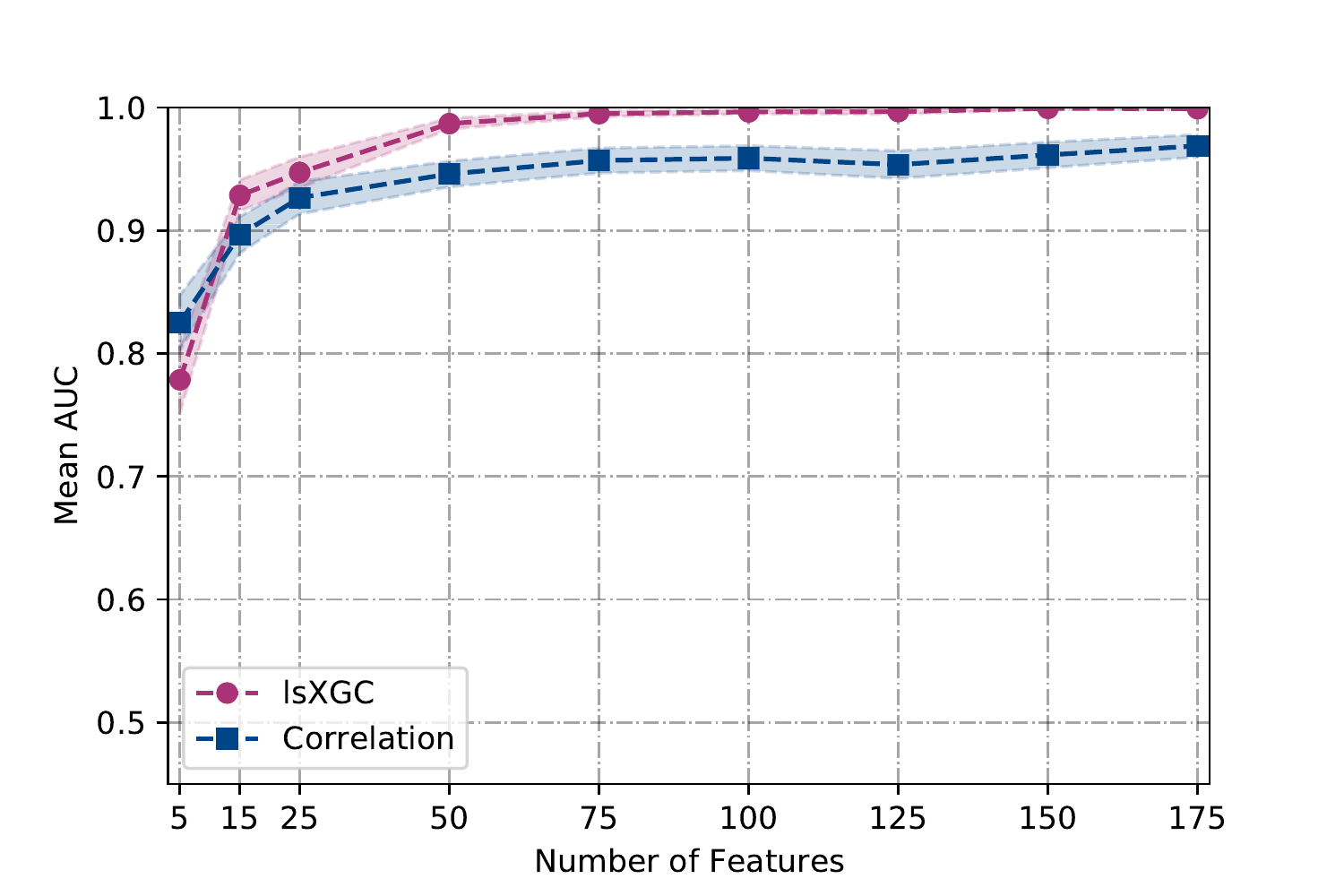}
         \caption{Mean AUC}
     \end{subfigure}
     \begin{subfigure}[b]{0.4\textwidth}
         \centering
         \includegraphics[width=\textwidth]{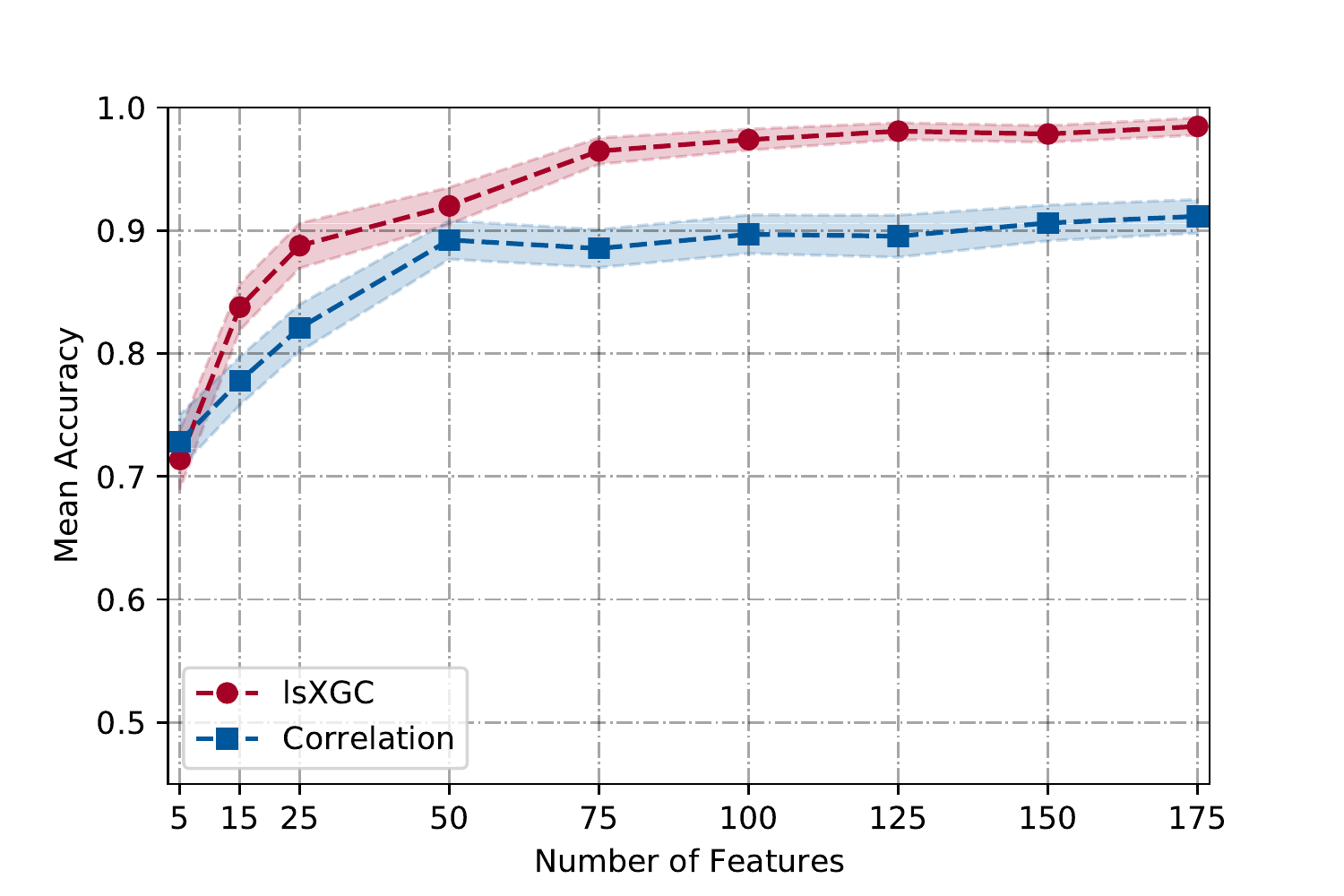}
         \caption{Mean accuracy}
     \end{subfigure}
     \vspace{.1in}
        \caption{Plots comparing the performance of cross-correlation and the proposed large-scale extended Granger causality (lsXGC). The shaded areas represent the $95\%$ confidence interval. It is clearly demonstrated that lsXGC outperforms cross-correlation for most numbers of selected features.}
        \label{fig:auc_plots}
\end{figure}
In this study, based on preliminary analyses, we chose 8 as the number of the retained components of PCA in the lsXGC algorithm, and model order of 4 for the multivariate vector autoregression function. From the plots of AUC and accuracy results in Fig. \ref{fig:auc_plots}, we can clearly see that lsXGC outperforms cross-correlation for varying numbers of features. Across the wide range of investigated numbers of features, the performance of lsXGC is consistently good with its mean AUC within [0.779, 0.999] and its mean accuracy within [0.714, 0.985]. On the other hand, cross-correlation performs quite poorly compared to lsXGC with its mean AUC within [0.825, 0.969] and its mean accuracy within [0.728, 0.912].

\section{CONCLUSIONS}\label{sec:conclusions}
In this study, we use a recently developed method for brain connectivity analysis, large-scale Extended Granger Causality (lsXGC), and apply it to subjects of the ACPI data repository in order to classify individuals history of the childhood ADHD diagnosis with frequent marijuana use from typical controls by estimating a measure of directed causal interaction between regional brain activities measured in resting-state fMRI. After constructing connectivity matrices as characterizing features for brain network analysis, we use Kendall's tau rank correlation coefficient for feature selection and a support vector machine for classification. We demonstrate that lsXGC compares favorably to conventional cross-correlation analysis, as indicated by the significantly higher AUC and accuracy values. The potential of lsXGC as an effective biomarker for identifying frequent marijuana use is yet to be validated in prospective clinical trials. However, our results suggest that our approach outperforms the current clinical standard, namely cross-correlation, at uncovering meaningful information from functional MRI data. 


\acknowledgments 
 
This research was funded by Ernest J. Del Monte Institute for Neuroscience Award from the Harry T. Mangurian Jr. Foundation. This work was conducted as a Practice Quality Improvement (PQI) project related to American Board of Radiology (ABR) Maintenance of Certificate (MOC) for Prof. Dr. Axel Wismüller. This work is not being and has not been submitted for publication or presentation elsewhere.  

\bibliography{report} 
\bibliographystyle{spiebib} 

\end{document}